# Research on Dynamic Data Flow Anomaly Detection based on Machine Learning

Liyang Wang[1*]
Olin Business School
Washington University
St. Louis, MO, 63130, USA
[*] Corresponding author e-mail address: liyang.wang@wustl.edu

Yu Cheng[2]
The Fu Foundation School of Engineering and Applied Science
Columbia University
New York, NY, 10027, USA
E-mail address: yucheng576@gmail.com

Hao Gong[3]
Independent Researcher
E-mail address: hgongscholar@gmail.com

Jiacheng Hu[4]
A. B. Freeman School of Business
Tulane University
New Orleans, LA, 70118, USA
E-mail address: jhu10@tulane.edu

Xirui Tang[5]
College of Computer Sciences
Northeastern University
Boston, MA, 02115, USA
E-mail address: tang.xir@northeastern.edu

Iris Li[6]
NYU Courant Institute of Mathematical Sciences
New York, NY 10012, USA
E-mail address: yl7580@nyu.edu

*Abstract*—The sophistication and diversity of contemporary cyberattacks have rendered the use of proxies, gateways, firewalls, and encrypted tunnels as a standalone defensive strategy inadequate. Consequently, the proactive identification of data anomalies has emerged as a prominent area of research within the field of data security. The majority of extant studies concentrate on sample equilibrium data, with the consequence that the detection effect is not optimal in the context of unbalanced data. In this study, the unsupervised learning method is employed to identify anomalies in dynamic data flows. Initially, multi-dimensional features are extracted from real-time data, and a clustering algorithm is utilised to analyse the patterns of the data. This enables the potential outliers to be automatically identified. By clustering similar data, the model is able to detect data behaviour that deviates significantly from normal traffic without the need for labelled data. The results of the experiments demonstrate that the proposed method exhibits high accuracy in the detection of anomalies across a range of scenarios. Notably, it demonstrates robust and adaptable performance, particularly in the context of unbalanced data.

## I. INTRODUCTION

The rapid development of information technology has led to the network becoming the infrastructure for social operations, thereby promoting the digital transformation of the global economy, finance, healthcare, education and other industries. However, the incidence of cyber attacks is also increasing, and the forms of attack are becoming more complex and diverse. These include distributed denial-of-service (DDoS) attacks, data theft, and malware, all of which represent a significant threat to network security [1]. The efficacy of traditional defensive mechanisms, such as firewalls, intrusion detection systems (IDS), and intrusion prevention systems (IPS), is contingent upon the availability of a comprehensive database of signatures and rules for known threats. These mechanisms are designed to be highly targeted. However, in the context of vast quantities of real-time data and the emergence of hitherto unknown threats, these traditional methods frequently appear to be insufficient and challenging to provide effective defences [2].

In particular, in an environment characterised by a dynamic data flow, data exhibits the attributes of real-time, continuous and massive data sets, which serves to exacerbate the complexity and velocity of change in network behaviour. Traditional static security protection mechanisms frequently prove inadequate in terms of detecting potential security threats in a timely manner. Furthermore, the expansion of network openness has led to a notable surge in the number of network nodes and terminal devices, accompanied by a substantial rise in data traffic, which has further compounded the challenge of identifying anomalous behaviour [3]. In these complex scenarios, traditional approaches that rely solely on rule matching and feature analysis are inadequate for detecting evolving cyber threats. Consequently, the ability to identify anomalous behaviours in a prompt and effective manner within a dynamic data flow has emerged as a pivotal challenge within the domain of cybersecurity research.

In recent years, machine learning techniques, particularly unsupervised learning, have been the subject of considerable interest within the field of cybersecurity. In contrast to conventional supervised learning techniques, which necessitate labelled data for training, unsupervised learning enables the automated identification of potential anomalies within data sets by examining their intrinsic structure and distribution patterns in the absence of known labels [4]. This renders unsupervised learning a highly promising avenue for addressing sophisticated and unanticipated attacks, particularly in the context of extensive, multidimensional data sets with limited labeling.

While existing research on machine learning-based anomaly detection has enhanced the detection capability for unknown threats to a degree, the majority of these studies remain constrained to the domain of sample equalisation. In other words, model training and evaluation are typically based on a substantial number of annotated normal and abnormal data sets, with the assumption that the distribution of the two types of samples is relatively balanced [5]. However, the actual network environment is often highly uneven, with the majority of traffic being normal and only a small number of abnormal behaviours. Furthermore, the distribution of anomalous traffic is frequently sparse and discrete, manifesting as a multitude of attack types, which further exacerbates the challenge of detection. Traditional anomaly detection models based on sample balancing often result in an elevated false positive rate or a high false negative rate when confronted with these imbalanced data sets, which is challenging to align with the demands of practical applications [6].

Additionally, the question of how to improve the accuracy of detection in scenarios where the data is unbalanced has become a topic of significant research interest. In order to address this challenge, this paper proposes a dynamic data flow anomaly detection model based on unsupervised learning. The model initially extracts multi-dimensional features from the real-time data set and then automatically groups the data set through the unsupervised clustering algorithm [7]. This enables the identification of abnormal traffic that is significantly different from the normal data set behaviour. In particular, the model categorises the majority of the normal data into a single category through the analysis of data similarity, and identifies potential abnormal behaviours by analysing the behavioural characteristics of a small number of deviant groups.

II. RELATED WORK

In order to address the prevalent anomaly issues in the Internet of Things (IoT), L. Nie et al. [8] integrated the pertinent insights from Convolutional Neural Networks (CNNs) to develop a CNN-based learning model. In contrast to conventional intrusion detection systems (IDS), the model is data-centric, with real-time data analysis for decision-making and action. The results of the experiments demonstrate that the proposed model is effective in detecting network intrusion, particularly in reducing the false positive rate. However, the low learning rate of the model results in a reduction in its learning efficiency when processing large-scale data, which limits its applicability in more complex scenarios. Nevertheless, the model continues to offer substantial advantages in enhancing network security protection capabilities, particularly in the context of evolving cyber threats. Its adaptability renders it a highly versatile tool with a vast range of potential applications.

Additionally, Fiore U et al. [9] employed a restricted Boltzmann machine (RBM) in conjunction with anomaly detection to ascertain that the RBM is capable of progressively discerning pivotal knowledge points from data through continuous learning. Gao N et al. further applied Deep Belief Networks (DBN) to anomaly detection, utilising DBN to learn a range of features within the data. The results demonstrate that DBN exhibits a robust capacity for discrimination when processing multi-dimensional data, effectively identifying anomalous behaviours. The results demonstrate that the integration of deep learning models and anomaly detection enhances the system's capacity to detect anomalies in complex data scenarios.

Furthermore, Farahnakian et al. [10] employed the capacity of autoencoders to diminish the dimensionality of multi-dimensional data, proposed the stacking of multiple autoencoders, and applied deep autoencoders to intrusion detection systems (IDS). The combination of greedy algorithms with unsupervised learning effectively circumvents the issue of model overfitting and the tendency to converge on local optimal solutions. This approach enhances the accuracy and robustness of intrusion detection, thereby improving its ability to cope with complex network environments.

III. METHODOLOGIES

A. Feature extraction

Initially, feature extraction represents a pivotal stage in the process of effective anomaly detection within a model. In the case of dynamic data flows, it is necessary to extract pertinent features from the raw data in order to represent changes in the data flow. The data stream $D(t)$ comprises $m$ samples, each comprising n-dimensional features represented as $X = \{x_1, x_2, ..., x_m\}$, where each $x_i \in \mathbb{R}^n$.

The data flow is continuous, so anomaly detection is not ideal for each data point directly. By introducing a time window $T$, we divide the data flow into several windows, each containing an $N_T$ number of data points, so that short-term fluctuations can be smoothed out and long-term trends can be preserved. After setting the time window $T$, the data in the window is $D_T = \{x_{t_1}, x_{t_2}, ..., x_{t_{N_T}}\}$, and we can calculate the statistical characteristics within each window, including mean $\mu_T = \frac{1}{N_T}\sum_{i=1}^{N_T} x_{t_i}$, variance $\sigma_T^2 = \frac{1}{N_T}\sum_{i=1}^{N_T}(x_{t_i} - \mu_T)^2$, and skewness $S_T = \frac{N_T}{(N_T-1)(N_T-2)}\sum_{i=1}^{N_T}(\frac{x_{t_i}-\mu_T}{\sigma_T})^3$.

In order to capture the periodic behavior and high-frequency changes in the data stream, a discrete Fourier transform is introduced to convert the time-domain signal into a frequency-domain signal. For each time window $D_T$, its Fourier transform is expressed as Equation 1.

$$X(f) = \sum_{t=0}^{N_T-1} x_t e^{\frac{j2\pi ft}{N_T}}, \#(1)$$

where $f$ is the frequency component. Using the Fourier transform, we can extract the major frequency components and power spectral density for each data window.

In order to prevent dimension disasters and feature redundancy, we further reduce the dimensionality of high-dimensional features by principal component analysis. The analysis uses the eigenvalue decomposition of the feature covariance matrix to find the direction of the maximum variance, denoted as Equation 2.

$$\Sigma = \frac{1}{m}\sum_{i=1}^{m}(x_i - \mu)(x_i - \mu)^T . \#(2)$$

Principal component analysis maps the original features to a new low-dimensional space, preserving the most important $d$ feature vectors $v_1, v_2, ..., v_d$, where $d$ makes the retained features explain more than 95% of the total variance.

Through the above steps, the final feature set contains both statistical features in the time domain and periodic features in the frequency domain, and is reduced by component analysis to improve the efficiency of the subsequent clustering algorithm.

### B. Unsupervised clustering algorithms

On the basis of feature extraction, this paper uses a density-based unsupervised clustering algorithm, Density Peaks Clustering (DPC), to cluster multi-dimensional features and automatically identify outliers. For each data point $x_i$, its local density $\rho_i$ is defined as Equation 3.

$$\rho_i = \sum_{j} \mathcal{X}(d(x_i - x_j) - d_c), \#(3)$$

where $d(x_i - x_j)$ is the distance between the points $x_i$ and $x_j$, which is obtained by Euclidean distance. Parameter $d_c$ is the truncation distance, which controls the range of density calculations, $\mathcal{X}(\cdot)$ is the indication function, $\mathcal{X}(\cdot) = 1$ when $x < 0$, otherwise 0. Local density $\rho_i$ represents the number of neighbor points within distance $d_c$, reflecting the degree of local aggregation of data points.

To distinguish between cluster centers of different densities, define the minimum distance $\delta_i$ from each data point to all points with higher densities, which is expressed as Equation 4.

$$\delta_i = \min_{j:\rho_j > \rho_i} d(x_i, x_j). \#(4)$$

For the point with the highest density, let $\delta_i = \max_j d(x_i, x_j)$. These high-density and space-flotted points are clustering centers. By sorting $\rho_i$ and $\delta_i$, the points with high local density and large distance are selected as the clustering centers. Each non-central point $x_i$ is assigned to the cluster with the higher density of the point $x_j$ closest to it.

Some of the significantly low-density and outlier points in the dataset can be directly marked as outliers. Define the anomaly score $A_i$ as Equation 5.

$$A_i = \frac{\delta_i}{\rho_i}. \#(5)$$

When $A_i$ exceeds a certain threshold of $A_{th}$, point $x_i$ is considered to be an anomaly. This approach eliminates the need for pre-labeling anomalies and automatically identifies potentially anomalous behavior from the data.

## IV. EXPERIMENTS

### A. Experimental setups

In this experiment, two network traffic datasets, NSL-KDD and UNSW-NB15, were used to construct an unsupervised anomaly detection model based on Density Peak Clustering (DPC) through multi-dimensional feature extraction (including time domain and frequency domain features) and PCA dimensionality reduction. In the experiment, traditional methods such as K-Means, Isolation Forest, and DBSCAN were compared.

Further, the proposed PCA is used to reduce the dimensionality of high-dimensional features to the first 50 principal components. In the clustering algorithm, the Density Peak Clustering algorithm sets the truncation distance $d_c = 0.15$, the local density threshold $\rho = 8$ and the distance threshold $\delta = 0.18$ to identify the clustering center and outliers. In addition, the threshold for anomaly score is set to $A_{th} = 1.4$, which is used to mark outliers as anomalous data.

### B. Experimental analysis

The accuracy of a model is defined as the proportion of correctly classified instances in the overall data set, including both normal and outlier cases. Figure 1 is a comparison of the accuracy of different methods, in which "Our Method" is marked with specific parameter settings, such as cut-off distance, local density threshold, distance threshold, and anomaly score threshold. With the different combinations of these parameters, three different configurations are shown in the diagram to show how our method performs with different parameter settings.

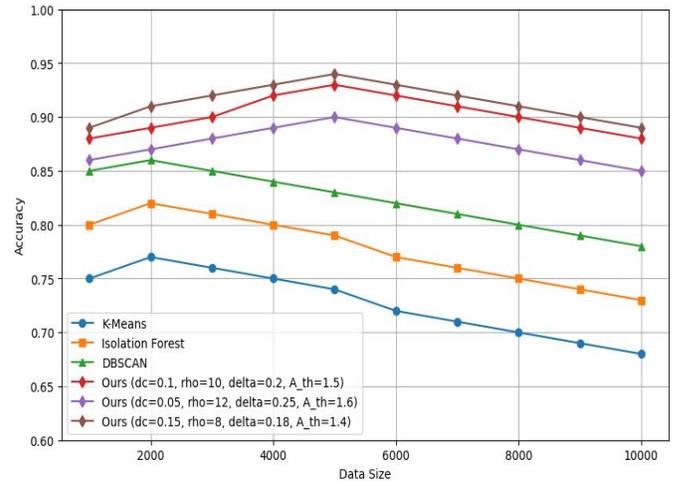

Figure 1. Accuracy Comparison of Different Methods with Ours

Furthermore, G-Mean is a measure of a model's performance on an unbalanced dataset that evaluates the model by balancing the classification performance of positive and negative classes.

Figure 2 shows the G-Mean comparison of different methods (K-Means, Isolation Forest, DBSCAN, and our proposed model with different parameter settings). The abscissa represents the amount of data, and the ordinate represents the G-Mean value, and the performance of each method at different data scales is clearly compared in a histogram.

The G-Mean values of K-Means, Isolation Forest, and DBSCAN fluctuate with the amount of data, but the performance of our method under different parameter configurations has always remained at a high level, especially for the parameter configurations of Settings 1 and 3, and the G-Mean is significantly better than the other methods. This shows that our model is more robust and stable when dealing with uneven data.

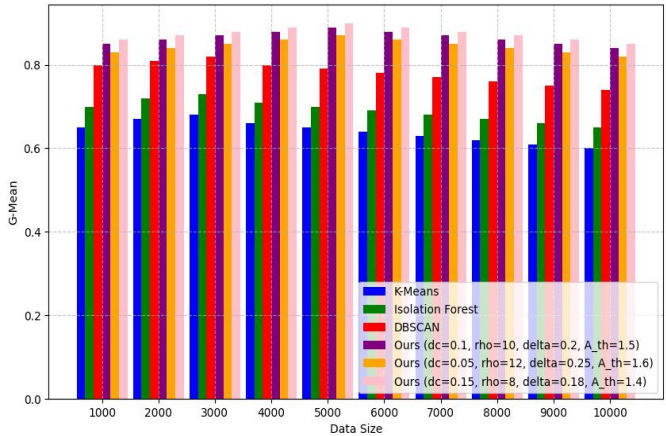

Figure 2.   G-Mean Comparison of Different Methods with Ours

The false positive rate is defined as the proportion of instances in which the model incorrectly identifies normal traffic as anomalous. The graph shows the comparison of the False Positive Rate (FPR) of different methods, including K-Means, Isolation Forest, DBSCAN, and the performance of our proposed model with different parameter settings. The abscissa represents the amount of data, and the ordinate represents the false positive rate.

As we can see in the Figure 3, the false positive rate of K-Means and Isolation Forest increases slightly as the amount of data increases, while the false positive rate of DBSCAN remains relatively low. Our model exhibits the lowest false positive rate at all three parameter settings, especially in parameter settings 1 and 3, where the false positive rate is significantly lower than that of other methods, demonstrating the effectiveness of the model in reducing false positives.

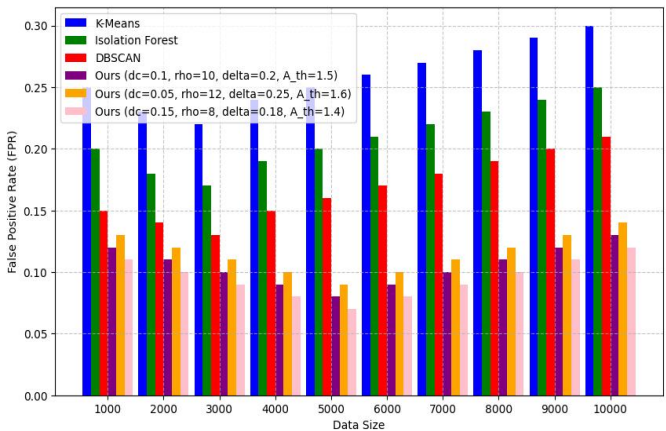

Figure 3.   False Positive Rate Comparison of Different Methods with Ours

## V. Conclusions

In conclusion, we proposed a dynamic data flow anomaly detection model based on unsupervised learning, which can effectively identify abnormal behaviors in network traffic through feature extraction, multi-dimensional clustering algorithm, and innovative anomaly score calculation. In the experiment, we used the NSL-KDD and UNSW-NB15 datasets, and compared them with classical methods such as K-Means, Isolation Forest, and DBSCAN, to evaluate the performance of accuracy, G-Mean, false positive rate and other indicators. Experimental results show that the proposed model has higher detection accuracy and robustness when dealing with unbalanced data, especially in reducing the false alarm rate. Through the testing of multiple parameter configurations, our method outperforms other methods in various scenarios, showing a wide range of application potential in the field of network security.